\documentclass[letterpaper]{article} % DO NOT CHANGE THIS
\usepackage{aaai2026}  % de-anonymized for arXiv preprint (no [submission] option)
\usepackage{times}  % DO NOT CHANGE THIS
\usepackage{helvet}  % DO NOT CHANGE THIS
\usepackage{courier}  % DO NOT CHANGE THIS
\usepackage[hyphens]{url}  % DO NOT CHANGE THIS
\usepackage{graphicx} % DO NOT CHANGE THIS
\usepackage{natbib}  % DO NOT CHANGE THIS AND DO NOT ADD ANY OPTIONS TO IT
\usepackage{caption} % DO NOT CHANGE THIS AND DO NOT ADD ANY OPTIONS TO IT
\usepackage{algorithm}
\usepackage{algorithmic}
\usepackage{float}        % enables [H] to force-place algorithm inline (no float, no gap)
\usepackage{booktabs}      % \toprule \midrule \bottomrule
\usepackage{multirow}      % table row spans
\usepackage{xspace}        % smart space after our macros
\usepackage{amsmath, amssymb, amsfonts}
\usepackage[table]{xcolor} % \columncolor / \cellcolor for table shading

\usepackage{newfloat}
\usepackage{listings}
\usepackage[most]{tcolorbox}
\usepackage[section]{placeins}  % \FloatBarrier to keep floats within sections
\DeclareCaptionStyle{ruled}{labelfont=normalfont,labelsep=colon,strut=off}
\floatstyle{ruled}
\newfloat{listing}{tb}{lst}{}
\floatname{listing}{Listing}

\nocopyright

\newcommand{\agora}{AGORA\xspace}

\newcommand{\mr}{\textit{mr}}
\newcommand{\Kr}{K_{\text{recent}}}

\newcommand{\nocomp}{\textsc{No-Comp}\xspace}
\newcommand{\obsmask}{\textsc{ObsMask}\xspace}
\newcommand{\longlingua}{\textsc{LongLLMLingua}\xspace}
\newcommand{\selctx}{\textsc{Selective Context}\xspace}

\newcommand{\agentdiet}{\textsc{AgentDiet}\xspace}
\newcommand{\acon}{\textsc{ACON}\xspace}

\newcommand{\alfworld}{\textsc{ALFWorld}\xspace}
\newcommand{\sciworld}{\textsc{ScienceWorld}\xspace}
\newcommand{\webshop}{\textsc{WebShop}\xspace}

\newcommand{\qwen}{\texttt{qwen3.5-flash}\xspace}
\newcommand{\gptmini}{\texttt{gpt-4o-mini}\xspace}
\newcommand{\gptfive}{\texttt{gpt-5-mini}\xspace}

\definecolor{exboxbody}{HTML}{F4F4F4}
\definecolor{exboxhead}{HTML}{B5B5B5}
\definecolor{exboxedge}{HTML}{C8C8C8}
\newtcolorbox{exbox}[1][Example]{%
  enhanced,
  breakable,
  colback=exboxbody,
  colframe=exboxhead,
  colbacktitle=exboxhead,
  coltitle=black,
  fonttitle=\sffamily\bfseries\small,
  fontupper=\small,
  arc=3pt,
  boxrule=0.6pt,
  left=7pt, right=7pt, top=4pt, bottom=4pt,
  toptitle=3pt, bottomtitle=3pt,
  title={#1},
}

\title{\agora: Adapter-Grounded Observation-Action Retention for\\Inference-Free Prompt Compression in LLM Agents}

\author{
    Haoran Zhang\textsuperscript{\rm 1},
    Zhaohua Sun\textsuperscript{\rm 2}\thanks{Corresponding author: joshsun@hku.hk}
}
\affiliations{
    \textsuperscript{\rm 1}AI Agent Technologies (Hong Kong) Limited\\
    \textsuperscript{\rm 2}Department of Mechanical Engineering, The University of Hong Kong\\
    8754817@gmail.com, joshsun@hku.hk
}

\begin{document}

\maketitle

\begin{abstract}
The token-level extractive compressors widely used for general LM context are structurally inappropriate for LLM agents: across $17$ (env, backbone, method) cells spanning two independent token-level method families, every cell collapses to mean reward $\mr{\leq}0.05$ despite $1.3$--$13.3\times$ realized compression. We name and characterize this failure mode as \emph{action-grammar destruction}---the tokens carrying action semantics (identifiers, brackets, action verbs) are exactly those self-information ranks lowest, so a general-purpose compressor reliably removes them and the environment rejects the residual. The diagnosis points to step-granularity compression. We introduce \agora{}, an inference-free step-level compressor combining a structural prompt parser, an always-keep floor for format- and recency-critical content, and a $125$M-parameter relevance scorer trained on counterfactual next-action-change labels (${\sim}2$ms/step, zero per-step LLM toll). Across the compared inference-free and LLM-based methods, \agora{} is the only one retaining $\geq 75\%$ uncompressed performance in $8$ of $9$ cells (with the lone exception at $73\%$); a four-way component ablation isolates the structural floor as the dominant quality lever and the learned scorer as the source of $1.0$--$11.5\times$ adaptive end-to-end compression from a single fixed keep ratio.
\end{abstract}

\begin{links}
    \link{Code}{https://github.com/ranranrannervous/agoracompression}
\end{links}

% ============================================================================
% 1. Introduction
% ============================================================================
\section{Introduction}
\label{sec:intro}
% !TEX root = ../agora.tex
% Section 1 — Introduction (paradigm-first; contributions as one flowing paragraph)

LLM agents accumulate trajectories that grow with task horizon: by step $30$ on \webshop{}, the in-context history can exceed $100$k tokens. Compressing it is a practical necessity, but compressing the prompt of an LLM agent is not structurally identical to compressing the prompt of a single-turn LM. Agents are notoriously sensitive to surface format: \textsc{AgentBench}~\citep{liu2024agentbench} names \emph{Invalid Format} and \emph{Invalid Action} as primary termination categories, meaning-preserving perturbations alone yield up to $76$ accuracy-point swings on ICL tasks~\citep{sclar2024quantifying}, and restricting LLM output format significantly degrades reasoning~\citep{tam2024let}. Compressing the prompt of a format-sensitive agent therefore carries a risk that compressing general LM context does not.

Two design families dominate. \emph{Token-level extractive} compressors~\citep{jiang2023llmlingua,jiang2024longllmlingua,li2023compressing,pan2024llmlingua} rank tokens by self-information or perplexity and drop those below a budget---cheap, and shown to deliver up to $20\times$ compression with minimal loss on QA and summarization. \emph{LLM-based abstractive} compressors invoke a secondary model per step (\textsc{HiAgent}~\citep{hu2025hiagent}, \textsc{ACON}~\citep{kang2025acon}) or per window (\textsc{AgentDiet}~\citep{xiao2025reducing})---preserving quality at the cost of $8$--$126$k extra tokens per task. The token-level family has not been systematically audited on agents. We do so, and find it fails not by a margin but \emph{structurally}.

Across $17$ (env, backbone, method) cells spanning \textsc{Selective Context}~\citep{li2023compressing} ($6$ cells) and \textsc{LLMLingua-2}~\citep{pan2024llmlingua} ($11$ cells, off-shelf plus a $4\times$-retrained variant), every cell collapses to $\mr{\leq}0.05$ despite $1.3$--$13.3\times$ realized compression. The mechanism is action-grammar destruction: tokens carrying action semantics---\texttt{search[men's shoes size 10]}, \texttt{click[Buy Now]}, identifiers, brackets, action verbs---are precisely those self-information ranks lowest because they are predictable from natural-text context. The compressor works as designed; the residual is grammatical English but ungrammatical agent commands the environment rejects. The two \textsc{LLMLingua-2} variants (off-shelf and retrained at the $4\times$ target on agent data) fail identically, ruling out hyperparameter remedies.

The diagnosis sharpens the design space: a compressor for agent prompts must operate at \emph{step granularity}, keeping each retained (action, observation) pair verbatim rather than pruning inside it. Existing step-level methods achieve this through a per-step LLM call; \citet{lindenbauer2025complexity} recently show that on coding agents, even a simple inference-free observation-masking rule matches LLM summarization, suggesting the inference-free corner deserves more attention. We push it further. We introduce \textbf{\agora{}} (Adapter-Grounded Observation-action Retention Architecture)---a structural parser that segments a prompt into role-typed blocks and (action, observation) step pairs, an always-keep floor pinning the system, task, current observation, and last $\Kr{=}2$ steps, and a $125$M-parameter RoBERTa relevance scorer trained on counterfactual next-action-change labels that allocates the remaining char budget; one ${\sim}2$ms forward pass per step, no API call. \agora{} is not proposed as a uniformly superior compressor: LLM-based methods undercut it on \$/task in $7$--$8$ of $9$ cells, and its cost advantage concentrates on long trajectories paired with expensive backbones.

Our contributions are as follows.

\noindent\textbf{A new failure mode for compression on agents.} We identify and name \emph{action-grammar destruction}---the structural reason token-level extractive compressors fail on agents, manifesting as collapse to $\mr{\leq}0.05$ in every cell of a $17$-cell audit across two independent token-level method families.

\noindent\textbf{\agora{}, an inference-free step-level compressor.} We propose a hierarchical compressor matching the diagnosis: a structural parser, an always-keep floor pinning format- and recency-critical content, and a $125$M-parameter relevance scorer trained on counterfactual next-action-change labels. \agora{} occupies the inference-free corner of step-level design space at ${\sim}2$ms per step and zero per-step LLM toll.

\noindent\textbf{Component attribution: which design choice buys what.} \agora{} is the only compared method retaining $\geq 75\%$ uncompressed performance in $8$ of $9$ cells (the lone exception at $73\%$). A four-way ablation (Table~\ref{tab:ablation-components}) isolates the always-keep floor as the dominant quality lever and the learned scorer as the source of adaptive end-to-end compression ($1.0$--$11.5\times$) from a single fixed keep ratio.

% ============================================================================
% 2. Related Work
% ============================================================================
\section{Related Work}
\label{sec:related}
% !TEX root = ../agora.tex
% Section 2 — Related Work (4 subsections, no standalone surveys block)

\subsection{Token-level Prompt Compression}

The dominant inference-free compressors rank individual tokens or sentences by an information criterion and drop the lowest-scoring units. \textsc{Selective Context}~\citep{li2023compressing} scores with self-information; \textsc{LLMLingua}~\citep{jiang2023llmlingua} and \textsc{LongLLMLingua}~\citep{jiang2024longllmlingua} use perplexity, the latter conditioned on the query; \textsc{LLMLingua-2}~\citep{pan2024llmlingua} reframes compression as token classification with a BERT-class encoder distilled from GPT-4; \textsc{RECOMP}~\citep{xu2024recomp} carries the same idea to RAG sentence selection with a downstream-task contrastive objective. The standard taxonomy of \citet{li2025prompt} groups these as \emph{hard-prompt} methods alongside the soft-prompt family discussed below, but its coverage is single-turn and does not currently distinguish a step-level agent-trajectory cell. All published evaluations are on QA, summarization, or long-context tasks; our \S\ref{sec:sensitivity} audit shows the rank-and-delete paradigm structurally fails on agents because the tokens that carry action semantics happen to be exactly those self-information ranks lowest. \agora{} departs from this family on three primitives: the unit of decision is a whole step (action+observation pair) rather than a token; the training signal is counterfactual next-action change rather than extractive faithfulness; and because each retained step is kept verbatim, brackets and identifiers inside action grammar are preserved by construction rather than by hoping a token scorer learns to.

\subsection{Soft-prompt and KV-cache Compression}

A second family modifies the LM rather than its input: gist tokens~\citep{mu2023learning}, in-context autoencoders~\citep{ge2023context}, and auto-compressors~\citep{chevalier2023adapting} learn soft prompts that summarize the input, while KV-cache eviction~\citep{xiao2024efficient,zhang2023h2o,li2024snapkv} compresses the attention cache directly. Both require white-box backbone access and produce non-natural-language artifacts, which puts them out of scope for our setting---frozen black-box agent backbones with no LM modification.

\subsection{Agent Context and Memory Management}

The closest neighbours to \agora{} compress agent trajectories step by step. \textsc{HiAgent}~\citep{hu2025hiagent} has the backbone formulate subgoals and then summarize prior observations within each subgoal scope; \textsc{ACON}~\citep{kang2025acon} compresses observations and interaction histories through a natural-language guideline iteratively optimized against failure traces; \textsc{AgentDiet}~\citep{xiao2025reducing} runs a GPT-5-mini reflection module over sliding windows on coding agents; \textsc{MemGPT}~\citep{packer2023memgpt} pages information between a fixed main context and external storage via LLM-issued function calls. All four invoke an LLM at compression time, paying the per-step token toll \agora{} avoids.

The closest \emph{inference-free} precedent is \citet{lindenbauer2025complexity}, who show on SWE-bench that a fixed observation-masking rule matches LLM summarization at roughly half the cost. Their rule is single-domain and has no learned component; \agora{} can be read as generalizing it---structural retention (role-block parser + $\Kr$-recent always-keep) plays the same quality-floor role, while the learned scorer extends realized compression beyond what that floor alone achieves. A separate body of work addresses cross-trial rather than within-trial memory---\textsc{A-Mem}~\citep{xu2026mem}, \textsc{Reflexion}~\citep{shinn2023reflexion}, \textsc{ExpeL}~\citep{zhao2024expel}, \textsc{Voyager}~\citep{wang2023voyager}---and is orthogonal to the working-memory compression problem we target. Using downstream-task utility as a training signal also has coarser-grained precedents in \textsc{UDR}~\citep{li2023unified}, \textsc{LLM-R}~\citep{wang2024learning}, and \textsc{RECOMP}~\citep{xu2024recomp}; \agora{}'s counterfactual labels (\emph{does removing this past step change the next action?}) specialize that idea to step-level agent context.

\subsection{Action-format Brittleness and Structural-over-semantic Findings}

Two background literatures sit behind our design. The first documents the format sensitivity of LMs and agents: format matters more than ground-truth labels in ICL~\citep{min2022rethinking}, meaning-preserving perturbations swing accuracy by up to $76$ points~\citep{sclar2024quantifying}, template choice can reduce strong models to random-guess level~\citep{voronov2024mind}, output-format restrictions degrade reasoning~\citep{tam2024let}, and position effects on retrieval are U-shaped~\citep{liu2024lost}---the latter directly motivating our $\Kr$ recency window. On the agent side, \textsc{AgentBench}~\citep{liu2024agentbench} names \emph{Invalid Format} and \emph{Invalid Action} as primary termination categories, and the multi-agent failure taxonomy of \citet{cemri2026multi} attributes $38.1\%$ of $1{,}242$ annotated traces to specification issues. The second is a structural-over-semantic pattern in LM context engineering: random documents can improve RAG accuracy by up to $35\%$ while semantically related-but-non-answer-bearing documents do the most damage~\citep{cuconasu2024power}, and multi-criteria reranking beats relevance-maximization alone~\citep{levine2025relevance}---both echoing the classical relevance-vs.-utility distinction from information retrieval~\citep{cooper1971definition,saracevic1975relevance}. \agora{} fits naturally into this pattern: token-level compressors optimize an informativeness axis that is mis-specified for agents, and the binding constraint turns out to be structural preservation of action grammar.

% ============================================================================
% 3. Method
% ============================================================================
\section{Method}
\label{sec:method}
% !TEX root = ../agora.tex
% Section 3 — Method (compact, eq fits 1-col).

% --- Figure 1 declared at the very top of Method so [!t] can promote it to
% the earliest possible top-of-page (one page earlier than placing it in §3.2).
\begin{figure*}[!t]
\centering
\includegraphics[width=\textwidth]{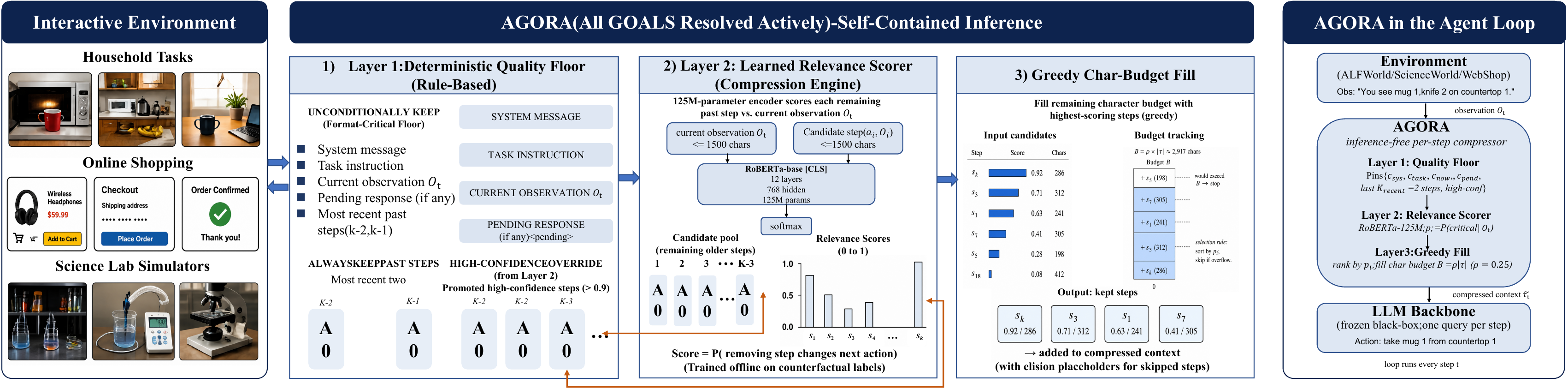}
\caption{Overview of \agora{}. At each step $t$, the agent's trajectory $\tau_t$ is parsed into role-typed blocks ($c_\text{sys}$: system, $c_\text{task}$: task, $\{s_i\}$: past (action, observation) steps, $c_\text{now}$: current observation, $c_\text{pend}$: pending assistant) and compressed by three sub-stages. \textbf{Layer 1 (Quality Floor)} unconditionally retains $\{c_\text{sys}, c_\text{task}, c_\text{now}, c_\text{pend}\}$, the last $\Kr{=}2$ steps, and any past step the scorer flags with $p_i{>}\theta_\text{hi}{=}0.9$. \textbf{Layer 2 (Relevance Scorer)} is a $125$M-parameter RoBERTa that outputs $p_i{=}P(\text{critical}\mid c_\text{now})$. \textbf{Layer 3 (Greedy Char-Budget Fill)} ranks remaining past steps by $p_i$ and fills the residual char budget $B{=}\rho\,|\tau_t|$ ($\rho{=}0.25$). The compressed context $\tilde\tau_t$ (target char budget $\rho{=}0.25$; realized end-to-end compression varies $1.0$--$11.5\times$ across cells, see \S\ref{sec:results-cost}) is then sent to the frozen backbone; the dashed return path marks the high-confidence override from Layer 2 back into Layer 1.}
\label{fig:overview}
\end{figure*}

\subsection{Problem Formulation}
\label{sec:method:problem}

At step $t$, an LLM agent has trajectory $\tau_t = (c_0, a_1, o_1, \ldots, a_{t-1}, o_{t-1}, o_t)$, where $c_0$ is the fixed system+task block and $(a_i, o_i)$ is the $i$-th action--observation pair. We insert a compressor $C$ between the environment and the backbone $\pi$, applied at every step:
\begin{equation}
\tilde\tau_t = C(\tau_t),\qquad |\tilde\tau_t| \,\leq\, \rho\,|\tau_t|,
\label{eq:compressor}
\end{equation}
with $|\cdot|$ measured in characters and $\rho \in (0,1]$. Two constraints distinguish agent compression from prompt compression in general: $C$ must be \emph{step-level}---surviving fragments must contain syntactically complete action calls, or the environment rejects them (\S\ref{sec:sensitivity})---and \emph{inference-free}, since the per-step LLM toll is precisely what we want to avoid.

\subsection{Approach Overview}
\label{sec:method:overview}

\agora{} comprises three components (Figure~\ref{fig:overview}). A \emph{structural parser} segments $\tau_t$ into role-typed blocks and groups them into past step pairs. An \emph{always-keep floor} unconditionally pins format-critical and recency-critical content. A learned \emph{step-relevance scorer} ranks the remaining past steps against the current observation, and a greedy procedure fills the residual char budget. Each retained step is kept verbatim, so action grammar is preserved by construction. The whole pass is a single $125$M-parameter forward---no API call, no LM modification, no white-box access to the agent backbone.

\subsection{Structural Parser and Always-Keep Floor}
\label{sec:method:floor}

Eval-time prompts follow the role-marker convention $\texttt{[SYSTEM]\ldots[USER]\ldots[ASSISTANT]\ldots}$. A single regex over role markers yields an ordered block list which a deterministic state machine groups into five role-typed components: a system block $c_\text{sys}$, the initial task instruction $c_\text{task}$, a sequence of past steps $\{s_i = (a_i, o_i)\}_{i=1}^{N}$ (each an $\texttt{[ASSISTANT]}+\texttt{[USER]}$ pair), the current observation $c_\text{now}$, and an optional pending $\texttt{[ASSISTANT]}$ block $c_\text{pend}$. The parser is task-agnostic at this level.

The first compression layer is a deterministic floor:
\begin{equation}
\begin{aligned}
\mathcal{K} \,=\,& \{c_\text{sys}, c_\text{task}, c_\text{now}, c_\text{pend}\} \,\cup\, \{s_i\}_{i=N-\Kr+1}^{N} \\
&\cup\, \{s_j : p_j > \theta_\text{hi}\},
\end{aligned}
\label{eq:floor}
\end{equation}
with $\Kr = 2$ and $\theta_\text{hi} = 0.9$. The three sets are, respectively, format-critical blocks whose removal consistently breaks rollouts; the $\Kr$ most recent steps, motivated by the U-shaped position effect of \citet{liu2024lost} and kept verbatim rather than at a token-budget rate; and a high-confidence override from the scorer below that lets the floor absorb past steps the relevance model is near-certain about, independent of recency. \S\ref{sec:ablation} (column $-$Scorer) shows that this floor alone---without the learned scorer---retains $\mr{=}0.356$ averaged over the $9$ cells ($\Delta\mr{=}{-}0.031$ vs full \agora{}-M22), so the floor carries the bulk of the quality retention and the scorer's incremental quality contribution is modest; the scorer's primary effect is on realized compression range (\S\ref{sec:results-cost}).

\subsection{Step Relevance Scorer and Greedy Budget Fill}
\label{sec:method:scorer}

A RoBERTa-base classifier $\phi$ (125M parameters) scores each past step against the current observation:
\begin{equation}
p_i \,=\, \big[\mathrm{softmax}(\phi(c_\text{now}^{\leq 1500},\;[a_i; o_i]^{\leq 1500}))\big]_{1},
\label{eq:scorer}
\end{equation}
where $(\cdot)^{\leq 1500}$ truncates each side at $1500$ characters to fit RoBERTa's $512$-subword window, and the subscript selects $P(\text{label}=\text{critical})$. The model is a single shared instance across all environments and backbones---no per-cell tuning.

Given budget $B = \lfloor \rho\,|\tau_t| \rfloor$ already partly spent by $\mathcal{K}$, Algorithm~\ref{alg:fill} fills the remainder by inserting non-floor past steps in descending $p_i$ order, skipping any that would exceed budget. Surviving steps are reinserted in chronological order; consecutive dropped steps are replaced by a single ``$[\ldots N\text{ step(s) elided}\ldots]$'' marker.

\begin{algorithm}[H]
\caption{Greedy char-budget fill (single step).}
\label{alg:fill}
\begin{algorithmic}[1]
\REQUIRE past steps $\{s_i\}_{i=1}^{N}$, scores $\{p_i\}$, floor $\mathcal{K}$, budget $B$
\ENSURE compressed trajectory $\tilde\tau_t$
\STATE $\mathcal{S} \gets \mathcal{K}$
\STATE $B_u \gets \sum_{x \in \mathcal{K}} |x|$
\STATE $\mathcal{R} \gets \{i : s_i \notin \mathcal{K}\}$ sorted by $p_i$ descending
\FOR{$i \in \mathcal{R}$}
  \IF{$B_u + |s_i| \leq B$}
    \STATE $\mathcal{S} \gets \mathcal{S} \cup \{s_i\}$
    \STATE $B_u \gets B_u + |s_i|$
  \ENDIF
\ENDFOR
\STATE assemble $\tilde\tau_t$ from $\mathcal{S}$ in chronological order
\RETURN $\tilde\tau_t$
\end{algorithmic}
\end{algorithm}

\subsection{Counterfactual Training}
\label{sec:method:training}

The scorer is trained on labels that operationalize \emph{``does this past step matter for the current decision?''} For each $(c_\text{now}, s_i)$ pair, we re-query the agent backbone $K{=}8$ times at sampling temperature $T{=}1.0$, half with the full trajectory and half with $s_i$ removed, canonicalize each sampled next action, and set
\begin{equation}
y_i \,=\, \frac{1}{K/2}\sum_{k=1}^{K/2}\mathbb{I}\!\left[\,\text{canon}(a^\star_k) \neq \text{canon}(a'_k)\,\right] \,\in\,[0,1],
\label{eq:cflabel}
\end{equation}
the fraction of paired rollouts whose canonicalized next action differs. Multi-rollout sampling ($K{=}8$) absorbs decoder stochasticity that a single-shot binary $y_i \in \{0,1\}$ would conflate with true criticality; \S\ref{sec:ablation} (column Hard-CF) shows replacing soft $y_i$ with the binary indicator costs $-0.059$ $\mr$ on average. Labels are generated on $21{,}523$ $(c_\text{now}, s_i)$ pairs from $1{,}244$ training trajectories spanning the three evaluation environments, with task initializations strictly disjoint from the evaluation tasks (Appendix~\ref{app:contamination}). The scorer is trained with a two-phase LP-FT schedule---$5$ epochs linear probing (backbone frozen, classification head only) followed by $4$ epochs full fine-tuning---under soft binary cross-entropy, AdamW, batch $16$, $10\%$ warmup + cosine decay; trajectory-level train/validation split ensures no past step appears in both partitions. Full hyperparameters are in Appendix~\ref{app:hparams}.

% ============================================================================
% 4. Experiments
% ============================================================================
\section{Experiments}
\label{sec:experiments}

% !TEX root = ../agora.tex
% Section 4.1 — Experimental Setup
% Style: top-conference (paragraph-led, rationale-integrated, no topic-then-list)

% --- Table 1 (Main Results) declared at the very top of §4 so [!t] can promote
% it to the page where §4 starts (one page earlier than placing it in §4.2).
\begin{table*}[!t]
\centering
\scriptsize
\setlength{\tabcolsep}{1pt}
\caption{Main results across all $9$ (env, backbone) cells. Top of each cell: $\mr$ (task-averaged reward, $n{=}30$) with retention $\mr_{\text{method}}/\mr_{\nocomp}$ in parentheses; bottom: $95\%$ bootstrap CI on $\mr$. \textbf{Bold} = best inference-free method per row; $^{\star}$ = retention ${>}100\%$; $^{\dagger}$ = per-step LLM call; shaded column = \agora{}. Summary block: Pass~$T$ rows count cells clearing retention threshold $T$ (Point). Eff.\,$\times$ range gives per-method min--max realized end-to-end compression; LongLingua's near-$1{\times}$ and Truncate's $\sim\!1.5{\times}$ are silent-compression failures (Table~\ref{tab:eff-ratio}).}
\label{tab:main}
\begin{tabular}{@{}llc>{\columncolor{gray!15}}cccccc|ccc@{}}
\toprule
 & & & \multicolumn{6}{c|}{Inference-free} & \multicolumn{3}{c}{LLM-call per step$^{\dagger}$} \\
\cmidrule(lr){4-9} \cmidrule(lr){10-12}
Env & Bb. & \nocomp & \agora & Trunc & Floor-K2 & ObsMask & LongLingua & SelCtx & HiAgent & AgentDiet & ACON \\
\midrule
\multirow{6}{*}{\alfworld}
  & \multirow{2}{*}{\qwen}
        & $.767$ ($-$) & $\mathbf{.767 (100)}$ & $\mathbf{.767 (100)}$  & $.600$ ($78$)  & $.300$ ($64$) & $.467$ ($100$) & $.033$ ($7$)  & $.400$ ($86$) & $.333$ ($71$) & $.400$ ($86$) \\
  &     & $[.60,.90]$  & $[.60,.90]$           & $[.60,.90]$            & $[.43,.77]$    & $[.13,.47]$   & $[.30,.63]$    & $[.00,.10]$   & $[.23,.57]$   & $[.17,.50]$   & $[.23,.57]$   \\
  & \multirow{2}{*}{\gptmini}
        & $.300$ ($-$) & $\mathbf{.333 (111)^{\star}}$ & $.267$ ($89$)   & $.233$ ($78$)  & $.100$ ($43$) & $.100$ ($43$)  & $.000$ ($0$)  & $.100$ ($43$) & $.167$ ($71$) & $.167$ ($71$) \\
  &     & $[.13,.47]$  & $[.17,.50]$           & $[.13,.43]$            & $[.10,.40]$    & $[.00,.23]$   & $[.00,.23]$    & $[.00,.00]$   & $[.00,.23]$   & $[.07,.30]$   & $[.03,.30]$   \\
  & \multirow{2}{*}{\gptfive}
        & $.767$ ($-$) & $\mathbf{.633 (83)}$  & $\mathbf{.633 (83)}$   & $.500$ ($65$)  & $.267$ ($53$) & $.467$ ($93$)  & $.000$ ($0$)  & $.400$ ($80$) & $.267$ ($53$) & $.233$ ($47$) \\
  &     & $[.60,.90]$  & $[.47,.80]$           & $[.47,.80]$            & $[.33,.67]$    & $[.13,.43]$   & $[.30,.63]$    & $[.00,.00]$   & $[.23,.57]$   & $[.13,.43]$   & $[.10,.40]$   \\
\midrule
\multirow{6}{*}{\sciworld}
  & \multirow{2}{*}{\qwen}
        & $.230$ ($-$) & $.210$ ($91$)         & $\mathbf{.285 (124)^{\star}}$ & $.147$ ($64$) & $.161$ ($82$) & $.201 (103)^{\star}$  & $.114$ ($58$) & $.164$ ($84$) & $.147$ ($75$) & $.146$ ($75$) \\
  &     & $[.15,.32]$  & $[.13,.31]$           & $[.18,.40]$            & $[.10,.20]$    & $[.10,.23]$   & $[.14,.28]$    & $[.07,.16]$   & $[.11,.23]$   & $[.08,.23]$   & $[.08,.23]$   \\
  & \multirow{2}{*}{\gptmini}
        & $.173$ ($-$) & $.155$ ($89$)         & $.156$ ($90$)          & $.164$ ($95$)  & $\mathbf{.154 (117)^{\star}}$ & $.144 (110)^{\star}$ & $.091$ ($69$) & $.146 (111)^{\star}$ & $.129$ ($98$) & $.132 (100)$ \\
  &     & $[.11,.25]$  & $[.11,.20]$           & $[.11,.21]$            & $[.11,.23]$    & $[.09,.23]$   & $[.09,.21]$    & $[.06,.13]$   & $[.08,.23]$   & $[.08,.19]$   & $[.08,.20]$   \\
  & \multirow{2}{*}{\gptfive}
        & $.298$ ($-$) & $.217$ ($73$)         & $\mathbf{.319 (107)^{\star}}$ & $.188$ ($63$) & $.215$ ($93$) & $.229$ ($99$)  & $.114$ ($49$) & $.206$ ($89$) & $.168$ ($72$) & $.172$ ($74$) \\
  &     & $[.21,.40]$  & $[.15,.30]$           & $[.22,.43]$            & $[.12,.27]$    & $[.14,.31]$   & $[.16,.31]$    & $[.07,.16]$   & $[.12,.31]$   & $[.10,.25]$   & $[.11,.25]$   \\
\midrule
\multirow{6}{*}{\webshop}
  & \multirow{2}{*}{\qwen}
        & $.499$ ($-$) & $.468$ ($94$)         & $.538 (108)^{\star}$   & $\mathbf{.561 (112)^{\star}}$ & $.594 (120)^{\star}$ & $.372$ ($75$) & $.000$ ($0$) & $.484$ ($98$) & $.567 (114)^{\star}$ & $.533 (107)^{\star}$ \\
  &     & $[.33,.67]$  & $[.30,.63]$           & $[.37,.70]$            & $[.41,.71]$    & $[.44,.75]$   & $[.23,.52]$    & $[.00,.00]$   & $[.33,.64]$   & $[.43,.71]$   & $[.37,.70]$   \\
  & \multirow{2}{*}{\gptmini}
        & $.293$ ($-$) & $.266$ ($91$)         & $.267$ ($91$)          & $\mathbf{.301 (103)^{\star}}$ & $.312 (223)^{\star}$ & $.113$ ($81$) & $.000$ ($0$) & $.278 (198)^{\star}$ & $.213 (152)^{\star}$ & $.227 (162)^{\star}$ \\
  &     & $[.16,.44]$  & $[.13,.41]$           & $[.15,.39]$            & $[.16,.45]$    & $[.20,.43]$   & $[.03,.21]$    & $[.00,.00]$   & $[.16,.41]$   & $[.09,.35]$   & $[.12,.35]$   \\
  & \multirow{2}{*}{\gptfive}
        & $.479$ ($-$) & $.431$ ($90$)         & $.359$ ($75$)          & $\mathbf{.509 (106)^{\star}}$ & $.507$ ($90$) & $.623 (111)^{\star}$ & $.000$ ($0$) & $.408$ ($73$) & $.637 (114)^{\star}$ & $.033$ ($6$)  \\
  &     & $[.32,.64]$  & $[.26,.60]$           & $[.21,.52]$            & $[.34,.68]$    & $[.35,.67]$   & $[.47,.77]$    & $[.00,.00]$   & $[.25,.57]$   & $[.49,.77]$   & $[.00,.10]$   \\
\midrule
\multicolumn{3}{@{}l}{Pass retention $\geq 75\%$ (Point)}      & $\mathbf{8/9}$ & $9/9$ & $6/9$ & $6/9$ & $8/9$ & $0/9$ & $7/9$ & $5/9$ & $5/9$ \\
\multicolumn{3}{@{}l}{Realized Eff.\,$\times$ range (9 cells)} & $1.0$--$11.5$ & $\sim\!1.5$ & $1.9$--$6.1$ & $1.3$--$4.8$ & $\mathbf{0.9-1.6}$ & $1.3$--$4.7$ & $2.4$--$8.3$ & $2.6$--$7.9$ & $3.0$--$14.0$ \\
\bottomrule
\end{tabular}
\end{table*}

\subsection{Experimental Setup}
\label{sec:setup}

\paragraph{Environments and backbones.}
We evaluate \agora{} on three text-based agent benchmarks---\textbf{ALFWorld}~\citep{shridhar2020alfworld}, \textbf{ScienceWorld}~\citep{wang2022scienceworld}, and \textbf{WebShop}~\citep{yao2022webshop}---paired with three production LLMs: \texttt{qwen3.5-flash}, \texttt{gpt-4o-mini}, and \texttt{gpt-5-mini}. This yields $3{\times}3{=}9$ (environment, backbone) cells; we hold out 30 evaluation tasks per environment, strictly disjoint from the $1{,}244$ trajectories used to generate \agora{}'s training labels (Appendix~\ref{app:contamination}).

\paragraph{Protocol.}
All rollouts use temperature $0$ and an episode budget of $30$ steps with fixed initialization seeds. Task initializations are shared across compression conditions within each cell. The \agora{} scorer is a single RoBERTa-base (125\,M parameters) classifier with $K_\text{recent}{=}2$ force-kept steps and target keep ratio $0.25$, held fixed across all 9 cells (no per-cell tuning). Training labels are derived from $K{=}8$ multi-rollout counterfactual sampling on the $1{,}244$ training trajectories; full label-generation protocol and hyperparameter values are in Appendix~\ref{app:hparams}.

\paragraph{Baselines.}
We compare \agora{} against ten methods, grouped by whether the compressor itself issues an LLM call. \emph{Inference-free}: \textbf{No-Comp} (uncompressed; quality ceiling), \textbf{Random-Step} (uniform random selection at the target keep ratio), \textbf{Truncate-2048} (hard last-$N$ token truncation), \textbf{Floor-K2} (structural heuristic that keeps only the last 2 past steps without learned scoring), \textbf{ObsMask}~\citep{lindenbauer2025complexity}, \textbf{LongLLMLingua}~\citep{jiang2024longllmlingua}, and \textbf{Selective Context}~\citep{li2023compressing}. \emph{LLM-based}: \textbf{HiAgent}~\citep{hu2025hiagent}, \textbf{AgentDiet}~\citep{xiao2025reducing}, and \textbf{ACON}~\citep{kang2025acon}. All budgeted methods target the same $4\times$ nominal compression ratio~\citep{jiang2023llmlingua}.

\paragraph{Metrics.}
We report four per-cell numbers: \emph{mean reward} ($\mr$, $n{=}30$), \emph{retention} ($\mr_\text{method}/\mr_\text{No-Comp}$), \emph{effective compression ratio} (end-to-end token reduction from API logs), and \emph{cost} (¥/task end-to-end, including any auxiliary compressor LLM calls). $95\%$ bootstrap CIs use $10$k resamples (Appendix~\ref{app:hparams}); significance is paired Wilcoxon signed-rank with Holm--Bonferroni correction.

% !TEX root = ../agora.tex
% Section 4.2 — Main Results: Quality Retention
% Full version: per-cell mr (retention) + 95% bootstrap CI on second line.
% Summary block: 5 rows (Pass thresholds at 75% point/CI-lower, 80/90 point, Eff× range).

% Table 2 (eff-ratio) declared here so [!t] can promote it onto an earlier page
% than its semantic owner §4.3 (Compression Ratio and Cost).
\begin{table*}[!htbp]
\centering
\scriptsize
\setlength{\tabcolsep}{4pt}
\caption{Realized compression ratio and per-task cost for the $5$ working methods across $9$ cells (LongLingua + SelCtx omitted as paradigm-fail, Table~\ref{tab:main}). Eff.\,$\times$: end-to-end token compression vs \nocomp{}. \$/task: per-task spend (USD), inclusive of any per-step compressor LLM tokens. \agora{} columns shaded. $^{\dagger}$: per-step LLM call. Per-method Eff.\,$\times$ ranges across the $9$ cells: \agora{} $1.0$--$11.5\times$, ObsMask $1.3$--$4.8\times$, HiAgent $2.4$--$8.3\times$, AgentDiet $2.6$--$7.9\times$, ACON $3.0$--$14.0\times$. LLM-based methods spend $8$--$126$k Meta-tokens per task on their own compressor call (inference-free methods spend $0$). LLM-based competitors beat \agora{} on \$/task in $8/9$ cells each; ObsMask wins $4/9$.}
\label{tab:eff-ratio}
\begin{tabular}{@{}ll >{\columncolor{gray!15}}cc cc cc cc cc@{}}
\toprule
 & & \multicolumn{2}{c}{\cellcolor{gray!15}\agora{}} & \multicolumn{2}{c}{ObsMask} & \multicolumn{2}{c}{HiAgent$^{\dagger}$} & \multicolumn{2}{c}{AgentDiet$^{\dagger}$} & \multicolumn{2}{c}{ACON$^{\dagger}$} \\
\cmidrule(lr){3-4} \cmidrule(lr){5-6} \cmidrule(lr){7-8} \cmidrule(lr){9-10} \cmidrule(lr){11-12}
Env & Backbone & Eff.\,$\times$ & \$/task & Eff.\,$\times$ & \$/task & Eff.\,$\times$ & \$/task & Eff.\,$\times$ & \$/task & Eff.\,$\times$ & \$/task \\
\midrule
\multirow{3}{*}{\alfworld}
  & \qwen     & $1.0$ & $.0010$ & $2.9$ & $.0004$ & $2.4$ & $.0005$ & $2.8$ & $.0004$ & $3.3$ & $.0003$ \\
  & \gptmini  & $1.7$ & $.0057$ & $3.6$ & $.0016$ & $3.9$ & $.0015$ & $7.8$ & $.0007$ & $8.2$ & $.0007$ \\
  & \gptfive  & $1.4$ & $.0105$ & $2.8$ & $.0048$ & $4.2$ & $.0029$ & $2.6$ & $.0052$ & $4.0$ & $.0034$ \\
\midrule
\multirow{3}{*}{\sciworld}
  & \qwen     & $1.8$ & $.0006$ & $1.3$ & $.0006$ & $2.9$ & $.0003$ & $3.6$ & $.0002$ & $3.2$ & $.0002$ \\
  & \gptmini  & $1.4$ & $.0019$ & $1.3$ & $.0023$ & $2.9$ & $.0010$ & $3.3$ & $.0009$ & $3.0$ & $.0010$ \\
  & \gptfive  & $\mathbf{11.5}$ & $.0160$ & $1.5$ & $.0146$ & $3.0$ & $.0082$ & $3.2$ & $.0082$ & $3.4$ & $.0077$ \\
\midrule
\multirow{3}{*}{\webshop}
  & \qwen     & $4.4$ & $.0026$ & $2.7$ & $.0018$ & $5.9$ & $.0010$ & $7.7$ & $.0008$ & $9.3$ & $.0008$ \\
  & \gptmini  & $8.9$ & $.0115$ & $4.8$ & $.0053$ & $8.3$ & $.0032$ & $7.9$ & $.0034$ & $\mathbf{14.0}$ & $.0021$ \\
  & \gptfive  & $3.3$ & $.0184$ & $2.2$ & $.0118$ & $3.8$ & $.0084$ & $5.4$ & $.0063$ & $4.4$ & $.0110$ \\
\bottomrule
\end{tabular}
\end{table*}

% Table 3 (ablation-components) declared here so [!t] can promote it onto an earlier
% page than its semantic owner §4.4 (Component Ablation).
\begin{table*}[!htbp]
\centering
\scriptsize
\setlength{\tabcolsep}{2pt}
\caption{Component ablation across all $9$ cells. Each variant disables one design choice in \agora{}-M22 and reuses the same deploy protocol ($n{=}30$ tasks/cell, $\rho{=}0.25$, $\Kr{=}2$ when applicable, identical task initializations, seed $47$). Top of each cell: $\mr$; bottom: $95\%$ bootstrap CI. \textbf{Bold}: best inference-free ablation per row. $^{\ast}$: paired Wilcoxon vs \agora{}-M22 has $p{<}0.05$ on the $30$ paired tasks. Shaded column = \agora{}-M22 (the deployed model). Mean and $\Delta$ rows aggregate across the $9$ cells.}
\label{tab:ablation-components}
\begin{tabular}{@{}llc>{\columncolor{gray!15}}ccccc|c@{}}
\toprule
 & & & & \multicolumn{4}{c|}{Component ablations} & Token-level \\
\cmidrule(lr){5-8} \cmidrule(lr){9-9}
Env & Bb. & \nocomp & \agora{}-M22 & $-$Floor & Hard-CF & $-$Scorer & Rand-Step & Trunc-2048 \\
\midrule
\multirow{6}{*}{\alfworld}
  & \multirow{2}{*}{\qwen}
        & $.767$        & $\mathbf{.767}$              & $.533^{\ast}$  & $.533^{\ast}$  & $.600^{\ast}$ & $.000^{\ast}$  & $\mathbf{.767}$ \\
  &     & $[.60,.90]$   & $[.60,.90]$                  & $[.37,.70]$    & $[.37,.70]$    & $[.43,.77]$   & $[.00,.00]$    & $[.60,.90]$ \\
  & \multirow{2}{*}{\gptmini}
        & $.300$        & $\mathbf{.333}$              & $.300$         & $.233$         & $.233$        & $.000^{\ast}$  & $.267$ \\
  &     & $[.13,.47]$   & $[.17,.50]$                  & $[.13,.47]$    & $[.10,.40]$    & $[.10,.40]$   & $[.00,.00]$    & $[.13,.43]$ \\
  & \multirow{2}{*}{\gptfive}
        & $.767$        & $\mathbf{.633}$              & $.400^{\ast}$  & $.467$         & $.500^{\ast}$ & $.000^{\ast}$  & $\mathbf{.633}$ \\
  &     & $[.60,.90]$   & $[.47,.80]$                  & $[.23,.57]$    & $[.30,.63]$    & $[.33,.67]$   & $[.00,.00]$    & $[.47,.80]$ \\
\midrule
\multirow{6}{*}{\sciworld}
  & \multirow{2}{*}{\qwen}
        & $.230$        & $\mathbf{.210}$              & $.113^{\ast}$  & $.162$         & $.147^{\ast}$ & $.109^{\ast}$  & $.285$ \\
  &     & $[.15,.32]$   & $[.13,.31]$                  & $[.08,.15]$    & $[.12,.21]$    & $[.10,.20]$   & $[.07,.15]$    & $[.18,.40]$ \\
  & \multirow{2}{*}{\gptmini}
        & $.173$        & $.155$                       & $.153$         & $.153$         & $\mathbf{.164}$ & $.092^{\ast}$ & $.156$ \\
  &     & $[.11,.25]$   & $[.11,.20]$                  & $[.11,.21]$    & $[.11,.21]$    & $[.11,.23]$   & $[.06,.12]$    & $[.11,.21]$ \\
  & \multirow{2}{*}{\gptfive}
        & $.298$        & $.217$                       & $.200$         & $\mathbf{.219}$ & $.188$       & $.105^{\ast}$  & $.319^{\ast}$ \\
  &     & $[.21,.40]$   & $[.15,.30]$                  & $[.12,.29]$    & $[.14,.32]$    & $[.12,.27]$   & $[.07,.15]$    & $[.22,.43]$ \\
\midrule
\multirow{6}{*}{\webshop}
  & \multirow{2}{*}{\qwen}
        & $.499$        & $.468$                       & $.255^{\ast}$  & $.440$         & $\mathbf{.561}$ & $.078^{\ast}$ & $.538$ \\
  &     & $[.33,.67]$   & $[.30,.63]$                  & $[.12,.40]$    & $[.28,.61]$    & $[.41,.71]$   & $[.00,.18]$    & $[.37,.70]$ \\
  & \multirow{2}{*}{\gptmini}
        & $.293$        & $.266$                       & $.286$         & $.299$         & $\mathbf{.301}$ & $.000^{\ast}$ & $.267$ \\
  &     & $[.16,.44]$   & $[.13,.41]$                  & $[.16,.41]$    & $[.16,.45]$    & $[.16,.45]$   & $[.00,.00]$    & $[.15,.39]$ \\
  & \multirow{2}{*}{\gptfive}
        & $.479$        & $.431$                       & $.450$         & $.442$         & $\mathbf{.509}$ & $.000^{\ast}$ & $.359$ \\
  &     & $[.32,.64]$   & $[.26,.60]$                  & $[.29,.62]$    & $[.28,.61]$    & $[.34,.68]$   & $[.00,.00]$    & $[.21,.52]$ \\
\midrule
\multicolumn{3}{@{}l}{Mean across $9$ cells}                       & $.387$ & $.299$ & $.328$ & $.356$ & $.043$ & $.399$ \\
\multicolumn{3}{@{}l}{$\Delta\mr$ vs \agora{}-M22}                 & ---    & $-.088$ & $-.059$ & $-.031$ & $-.344$ & $+.012$ \\
\multicolumn{3}{@{}l}{Cells with $p{<}0.05$ vs \agora{}-M22 ($/9$)} & --- & $\mathbf{4}$ & $1$ & $3$ & $\mathbf{9}$ & $1$ \\
\bottomrule
\end{tabular}
\end{table*}

\subsection{Main Results: Quality Retention}
\label{sec:results-main}

% Table 1 (\ref{tab:main}) is declared at the top of 4_1_experimental_setup.tex
% so it lands on the page where §4 starts (one page earlier).

\begin{figure*}[!t]
\centering
\includegraphics[width=0.95\textwidth]{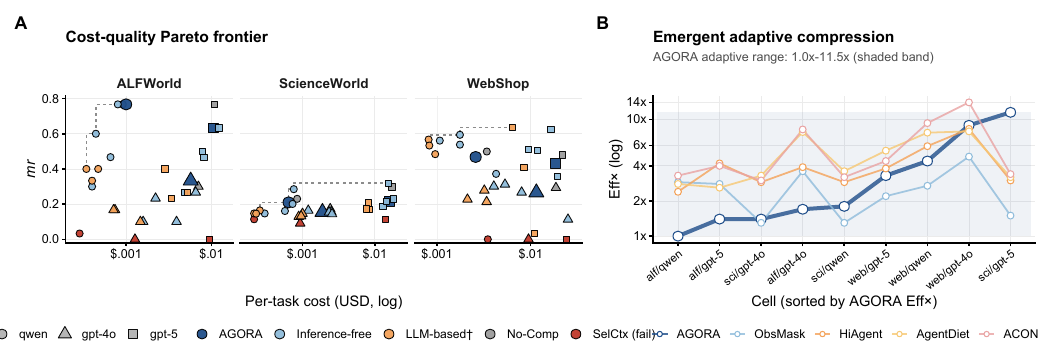}
\caption{\textbf{(A)}~Cost--quality Pareto frontier across the $9$ cells (per-env facets; dashed = per-env Pareto, excluding \nocomp{} and \selctx{}). \agora{} (deep blue) leads on the inference-free frontier at the $4{\times}$ operating point, dominating Floor-K2 at matched compression and Pareto-improving over Truncate-2048 (which only realizes ${\sim}1.5{\times}$). \textbf{(B)}~Realized end-to-end Eff.\,$\times$ per method (log $y$, ordered by \agora{}'s ratio). The shaded band is \agora{}'s adaptive $1.0$--$11.5\times$ end-to-end range emerging from a single fixed scorer keep-ratio $\rho{=}0.25$.}
\label{fig:main-composite}
\end{figure*}

At the $4{\times}$ compression target, \agora{} achieves average $\mr{=}0.387$ ($\approx 92\%$ retention vs \nocomp{}), outperforming structural Floor-K2 ($\mr{=}0.356$, $85\%$) by $+7$pp at matched compression. Truncate-2048's apparently higher $\mr{=}0.399$ reflects only $\approx 1.5{\times}$ realized compression (\S\ref{sec:results-cost})---not Pareto-comparable at the $4{\times}$ operating point. \agora{} clears $\geq 75\%$ retention in $8/9$ cells (\sciworld{}$\times$\gptfive{} at $73\%$); no other inference-free method at $4{\times}$ achieves this. We focus on retention because the deployment question is how many cells a compressor keeps within $75\%$ at the target budget, not absolute improvement.

At $n{=}30$ tasks per cell, $95\%$ CIs are wide ($1$--$2$ cells clear any threshold by CI-lower across all methods); we therefore report point estimates.

The two token-level methods fail in opposite ways. \longlingua{} fails to compress: set to $4\times$ but realized at $0.9$--$1.6\times$, so its competitive retention reflects an essentially uncompressed input rather than a working method. \selctx{} fails to act: it does hit $1.3$--$4.7\times$ real compression, yet collapses the agent on every cell---token-level self-information selection is structurally incompatible with agent action grammar (e.g.\ dropping tokens from \texttt{search[men's shoes size 10]} yields ungrammatical \webshop{} actions). We replicate this paradigm failure across $17$ additional cells in \S\ref{sec:sensitivity}.

Figure~\ref{fig:main-composite}(A) places \agora{} on the inference-free Pareto frontier in every env facet; (B) shows that a single fixed $\rho{=}0.25$ produces a $1.0$--$11.5\times$ end-to-end range across cells, a region LLM-based methods reach only with per-step LLM overhead (Table~\ref{tab:eff-ratio}).

% !TEX root = ../agora.tex
% Section 4.3 — Compression Ratio and Cost
% (Table 2 declaration was moved to top of 4_2_main_results.tex to let float queue
% promote it onto the earlier page.)

\subsection{Compression Ratio and Cost}
\label{sec:results-cost}

Table~\ref{tab:eff-ratio} decomposes per-task cost into realized end-to-end Eff.\,$\times$ and \$/task (cost formula, pricing constants, and the per-task Eff.\,$\times$ definition are in Appendix~\ref{app:cost-accounting}). Although every budgeted method was targeted at the same nominal $4\times$, realized ratios diverge sharply: \agora{} spans $1.0$--$11.5\times$ with no method-side adjustment, adapting to the trajectory regime (peak on \sciworld{}$\times$\gptfive{} where past observations dominate input, bottom on the short \alfworld{}$\times$\qwen{} cell where the structural floor already covers most of the context). \acon{} covers a wider \emph{maximum} ($3.0$--$14.0\times$), but pays $14$--$94$k Meta-tokens per task for it; \obsmask{}'s position-only rule yields a narrower $1.3$--$4.8\times$. What distinguishes \agora{} is the combination---wide adaptive range \emph{at zero meta-overhead}---which we attribute in \S\ref{sec:ablation} to the learned scorer treating the keep-ratio as a soft prior rather than a hard budget.

The three LLM-based methods each beat \agora{} on \$/task in $7$--$8$ of $9$ cells: their per-step LLM toll is small relative to the savings from aggressive compression. \agora{} is therefore not a universal cost winner. Its differentiator is \emph{zero meta-token overhead}: every other competitive method consumes $8$--$126$k extra tokens per task on its own compressor LLM call, and \agora{} consumes zero. The deployments where this matters are exactly those where retention is the binding constraint and the per-step LLM toll cannot be absorbed---\alfworld{}$\times$\gptmini{} is illustrative: \agora{} actually \emph{exceeds} \nocomp{} quality ($\mr{=}0.333$ vs $0.300$, $111\%$ retention) at $1.7\times$ compression with zero compressor overhead, while every LLM-based competitor sits at $\mr{\leq}0.167$ on this cell. On the inverse end (long-trajectory webshop, where LLM-based aggressive compression yields large savings), \acon{} or \agentdiet{} are reasonable choices; our results should not be read as universal advocacy for inference-free compression. We return to this characterization in \S\ref{sec:discussion}.

% !TEX root = ../agora.tex
% Section 4.4 — Component Ablation
% (Table 3 declaration moved to top of 4_2_main_results.tex to let float queue
% promote it onto an earlier page.)

\subsection{Component Ablation}
\label{sec:ablation}

Table~\ref{tab:ablation-components} isolates the contribution of each design choice in \agora{}-M22 by disabling one at a time and re-deploying on the same $9$-cell, $30$-task grid (paired by task initialization, seed $47$). \textbf{$-$Floor} disables the $\Kr{=}2$ structural always-keep window; \textbf{Hard-CF} replaces the $K{=}8$ multi-rollout soft labels with single-shot binary ones; \textbf{$-$Scorer} keeps only the floor and the parser (no learned step relevance); \textbf{Rand-Step} replaces the scorer with uniform random scores at the same target $\rho{=}0.25$; \textbf{Trunc-2048} is a token-level baseline that ignores step structure entirely.

The decomposition shows a consistent hierarchy of contributions. \textbf{Removing the floor is the costliest single change} ($-0.088\,\mr$ on average, $4/9$ cells significant), because the floor underwrites the basic ``the last two steps are always present'' invariant that downstream action-selection assumes. \textbf{Multi-rollout soft labels matter more than the scorer's structural choice} ($-0.059$ for Hard-CF vs $-0.031$ for $-$Scorer): replacing $K{=}8$ soft labels with $K{=}1$ hard binary loses signal that single-shot sampling treats as label noise. \textbf{Removing the scorer entirely costs only $-0.031$ on average}: structurally, $-$Scorer is Floor-K2 (Table~\ref{tab:main}), so the scorer's contribution is primarily \emph{realized compression range} ($1.0$--$11.5\times$), not \mr. \textbf{Random scoring is catastrophic} ($-0.344$, $9/9$ cells significant), confirming that the scorer's signal---even at modest mean $\Delta\mr$---is the load-bearing component once the floor is in place.

Trunc-2048's apparent parity ($+0.012$ $\mr$) is bought with $\approx 1.5\times$ realized compression---far below the $4\times$ target and \agora{}-M22's $1.0$--$11.5\times$ range---so it is not Pareto-comparable here. The ablation hierarchy is sharpest on \alfworld{} and weakest on \webshop{}, where the structural floor is already so competitive that $-$Scorer occasionally improves $\mr$ (e.g.\ $0.561$ vs $0.468$ on \webshop{}$\times$\qwen, mirroring Table~\ref{tab:main} where Floor-K2 wins this cell)---the scorer's value concentrates on shorter trajectories where per-step relevance signal is strongest.

% !TEX root = ../agora.tex
% Section 4.5 — Paradigm failure of token-level compression (C1, hoisted to lead)

\subsection{Paradigm Failure of Token-Level Compression}
\label{sec:sensitivity}

Token-level extractive compression---the dominant paradigm for general LLM context reduction---is structurally inappropriate for agent settings. We provide the first systematic evidence across $17$ (environment, backbone, method) cells spanning two independent token-level method families. \selctx{}~\citep{li2023compressing} fails on all $6/6$ cells of our main grid. \textsc{LLMLingua-2}~\citep{jiang2024longllmlingua} fails on all $11/11$ cells of an extended audit (off-shelf with \texttt{rate=0.25} on $6$ cells, plus a $4\times$-retrained variant on $5$ cells; both variants collapse identically). Aggregated, $17/17$ cells collapse to $\mr{\leq}0.05$ despite achieving genuine $1.3$--$13.3\times$ realized compression. The compressor works as advertised at the token level; the agent does not.

The failure mechanism is action-grammar destruction. Token-level methods rank by self-information or perplexity, but the tokens that carry action semantics in agent trajectories---item identifiers, bracket delimiters, action verbs, prepositional connectors---are precisely the tokens that language-model statistics score lowest, because they are highly predictable from natural-text context. The compressor reliably removes them; the residual text is grammatical English but ungrammatical agent commands such as \texttt{search men's shoes 10} or \texttt{click[]}, which the environment rejects regardless of task content. This rules out hyperparameter remedies (both \textsc{LLMLingua-2} variants fail identically) and task-specific artifacts (failure is task-independent; Appendix~\ref{app:contamination} shows no phase-B contamination on this audit). The per-cell breakdown of all $17$ cells is given in Appendix~\ref{app:paradigm-detail}.

This paradigm-level diagnosis motivates the design space \agora{} occupies: \emph{step-level} compression (operating on intact actions and observations rather than tokens), with an \emph{inference-free} implementation (no per-step LLM call). Within that design space, \agora{}'s two budget knobs ($\rho{=}0.25$, $K_\text{recent}{=}2$) were fixed across all $9$ cells before evaluation, inherited from training-time defaults; no per-cell tuning was performed, so the cost--retention frontier reported in Table~\ref{tab:eff-ratio} is the one practitioners see at deploy time.

% \input{sections/4_6_mechanism}  % removed: ObsMask had no deploy data, 3 diagnostic cells reversed/unsupported

% ============================================================================
% 5. Conclusion (absorbs Discussion; sec:discussion label kept for back-refs)
% ============================================================================
\section{Conclusion}
\label{sec:conclusion}
\label{sec:discussion}
% !TEX root = ../agora.tex
% Section 5 — Conclusion. Single short paragraph + bolded run-in Limitations.

This paper diagnoses why general token-level prompt compression fails on LLM agents (\emph{action-grammar destruction}: tokens carrying action semantics are exactly those self-information ranks lowest) and proposes \agora{}, an inference-free step-level alternative whose quality guarantee sits on a structural floor and whose learned scorer supplies adaptive end-to-end compression---an attribution we expect to generalize to richer agent settings where action grammars are stricter still.

\noindent\textbf{Limitations.} $n{=}30$ per cell gives wide $95\%$ CIs (point estimates only); criticality is partly backbone-specific.

% ============================================================================
% Acknowledgments (unnumbered, before references)
% ============================================================================
\section*{Acknowledgments}
This work was funded by AI Agent Technologies (Hong Kong) Limited.

% ============================================================================
% Bibliography
% ============================================================================
% aaai2026.sty already sets \bibliographystyle, do not redeclare here.
\bibliography{aaai2026}

\begin{thebibliography}{37}
\providecommand{\natexlab}[1]{#1}

\bibitem[{Cemri et~al.(2026)Cemri, Pan, Yang, Agrawal, Chopra, Tiwari, Keutzer,
  Parameswaran, Klein, Ramchandran et~al.}]{cemri2026multi}
Cemri, M.; Pan, M.~Z.; Yang, S.; Agrawal, L.~A.; Chopra, B.; Tiwari, R.;
  Keutzer, K.; Parameswaran, A.; Klein, D.; Ramchandran, K.; et~al. 2026.
\newblock Why do multi-agent llm systems fail?
\newblock \emph{Advances in Neural Information Processing Systems}, 38.

\bibitem[{Chevalier et~al.(2023)Chevalier, Wettig, Ajith, and
  Chen}]{chevalier2023adapting}
Chevalier, A.; Wettig, A.; Ajith, A.; and Chen, D. 2023.
\newblock Adapting language models to compress contexts.
\newblock In \emph{Proceedings of the 2023 Conference on Empirical Methods in
  Natural Language Processing}, 3829--3846.

\bibitem[{Cooper(1971)}]{cooper1971definition}
Cooper, W.~S. 1971.
\newblock A definition of relevance for information retrieval.
\newblock \emph{Information storage and retrieval}, 7(1): 19--37.

\bibitem[{Cuconasu et~al.(2024)Cuconasu, Trappolini, Siciliano, Filice,
  Campagnano, Maarek, Tonellotto, and Silvestri}]{cuconasu2024power}
Cuconasu, F.; Trappolini, G.; Siciliano, F.; Filice, S.; Campagnano, C.;
  Maarek, Y.; Tonellotto, N.; and Silvestri, F. 2024.
\newblock The power of noise: Redefining retrieval for rag systems.
\newblock In \emph{Proceedings of the 47th International ACM SIGIR Conference
  on Research and Development in Information Retrieval}, 719--729.

\bibitem[{Ge et~al.(2023)Ge, Hu, Wang, Wang, Chen, and Wei}]{ge2023context}
Ge, T.; Hu, J.; Wang, L.; Wang, X.; Chen, S.-Q.; and Wei, F. 2023.
\newblock In-context autoencoder for context compression in a large language
  model.
\newblock \emph{arXiv preprint arXiv:2307.06945}.

\bibitem[{Hu et~al.(2025)Hu, Chen, Chen, Mu, Shao, and Luo}]{hu2025hiagent}
Hu, M.; Chen, T.; Chen, Q.; Mu, Y.; Shao, W.; and Luo, P. 2025.
\newblock Hiagent: Hierarchical working memory management for solving
  long-horizon agent tasks with large language model.
\newblock In \emph{Proceedings of the 63rd Annual Meeting of the Association
  for Computational Linguistics (Volume 1: Long Papers)}, 32779--32798.

\bibitem[{Jiang et~al.(2023)Jiang, Wu, Lin, Yang, and Qiu}]{jiang2023llmlingua}
Jiang, H.; Wu, Q.; Lin, C.-Y.; Yang, Y.; and Qiu, L. 2023.
\newblock Llmlingua: Compressing prompts for accelerated inference of large
  language models.
\newblock In \emph{Proceedings of the 2023 conference on empirical methods in
  natural language processing}, 13358--13376.

\bibitem[{Jiang et~al.(2024)Jiang, Wu, Luo, Li, Lin, Yang, and
  Qiu}]{jiang2024longllmlingua}
Jiang, H.; Wu, Q.; Luo, X.; Li, D.; Lin, C.-Y.; Yang, Y.; and Qiu, L. 2024.
\newblock Longllmlingua: Accelerating and enhancing llms in long context
  scenarios via prompt compression.
\newblock In \emph{Proceedings of the 62nd Annual Meeting of the Association
  for Computational Linguistics (Volume 1: Long Papers)}, 1658--1677.

\bibitem[{Kang et~al.(2025)Kang, Chen, Han, Inan, Wutschitz, Chen, Sim, and
  Rajmohan}]{kang2025acon}
Kang, M.; Chen, W.-N.; Han, D.; Inan, H.~A.; Wutschitz, L.; Chen, Y.; Sim, R.;
  and Rajmohan, S. 2025.
\newblock Acon: Optimizing context compression for long-horizon llm agents.
\newblock \emph{arXiv preprint arXiv:2510.00615}.

\bibitem[{LeVine and Varjavand(2025)}]{levine2025relevance}
LeVine, W.; and Varjavand, B. 2025.
\newblock Relevance Isn't All You Need: Scaling RAG Systems With Inference-Time
  Compute Via Multi-Criteria Reranking.
\newblock \emph{arXiv preprint arXiv:2504.07104}.

\bibitem[{Li et~al.(2023{\natexlab{a}})Li, Lv, Yan, Lin, Zhu, Ni, Xie, Wang,
  and Qiu}]{li2023unified}
Li, X.; Lv, K.; Yan, H.; Lin, T.; Zhu, W.; Ni, Y.; Xie, G.; Wang, X.; and Qiu,
  X. 2023{\natexlab{a}}.
\newblock Unified demonstration retriever for in-context learning.
\newblock In \emph{Proceedings of the 61st Annual Meeting of the Association
  for Computational Linguistics (Volume 1: Long Papers)}, 4644--4668.

\bibitem[{Li et~al.(2023{\natexlab{b}})Li, Dong, Guerin, and
  Lin}]{li2023compressing}
Li, Y.; Dong, B.; Guerin, F.; and Lin, C. 2023{\natexlab{b}}.
\newblock Compressing context to enhance inference efficiency of large language
  models.
\newblock In \emph{Proceedings of the 2023 conference on empirical methods in
  natural language processing}, 6342--6353.

\bibitem[{Li et~al.(2024)Li, Huang, Yang, Venkitesh, Locatelli, Ye, Cai, Lewis,
  and Chen}]{li2024snapkv}
Li, Y.; Huang, Y.; Yang, B.; Venkitesh, B.; Locatelli, A.; Ye, H.; Cai, T.;
  Lewis, P.; and Chen, D. 2024.
\newblock Snapkv: Llm knows what you are looking for before generation.
\newblock \emph{Advances in Neural Information Processing Systems}, 37:
  22947--22970.

\bibitem[{Li et~al.(2025)Li, Liu, Su, and Collier}]{li2025prompt}
Li, Z.; Liu, Y.; Su, Y.; and Collier, N. 2025.
\newblock Prompt compression for large language models: A survey.
\newblock In \emph{Proceedings of the 2025 Conference of the Nations of the
  Americas Chapter of the Association for Computational Linguistics: Human
  Language Technologies (Volume 1: Long Papers)}, 7182--7195.

\bibitem[{Lindenbauer et~al.(2025)Lindenbauer, Slinko, Felder, Bogomolov, and
  Zharov}]{lindenbauer2025complexity}
Lindenbauer, T.; Slinko, I.; Felder, L.; Bogomolov, E.; and Zharov, Y. 2025.
\newblock The Complexity Trap: Simple Observation Masking Is as Efficient as
  LLM Summarization for Agent Context Management.
\newblock \emph{arXiv preprint arXiv:2508.21433}.

\bibitem[{Liu et~al.(2024{\natexlab{a}})Liu, Lin, Hewitt, Paranjape,
  Bevilacqua, Petroni, and Liang}]{liu2024lost}
Liu, N.~F.; Lin, K.; Hewitt, J.; Paranjape, A.; Bevilacqua, M.; Petroni, F.;
  and Liang, P. 2024{\natexlab{a}}.
\newblock Lost in the middle: How language models use long contexts.
\newblock \emph{Transactions of the association for computational linguistics},
  12: 157--173.

\bibitem[{Liu et~al.(2024{\natexlab{b}})Liu, Yu, Zhang, Xu, Lei, Lai, Gu, Ding,
  Men, Yang et~al.}]{liu2024agentbench}
Liu, X.; Yu, H.; Zhang, H.; Xu, Y.; Lei, X.; Lai, H.; Gu, Y.; Ding, H.; Men,
  K.; Yang, K.; et~al. 2024{\natexlab{b}}.
\newblock Agentbench: Evaluating llms as agents.
\newblock In \emph{International Conference on Learning Representations},
  volume 2024, 52989--53046.

\bibitem[{Min et~al.(2022)Min, Lyu, Holtzman, Artetxe, Lewis, Hajishirzi, and
  Zettlemoyer}]{min2022rethinking}
Min, S.; Lyu, X.; Holtzman, A.; Artetxe, M.; Lewis, M.; Hajishirzi, H.; and
  Zettlemoyer, L. 2022.
\newblock Rethinking the Role of Demonstrations: What Makes In-Context Learning
  Work?
\newblock In \emph{Proceedings of the 2022 Conference on Empirical Methods in
  Natural Language Processing}, 11048--11064.

\bibitem[{Mu, Li, and Goodman(2023)}]{mu2023learning}
Mu, J.; Li, X.; and Goodman, N. 2023.
\newblock Learning to compress prompts with gist tokens.
\newblock \emph{Advances in Neural Information Processing Systems}, 36:
  19327--19352.

\bibitem[{Packer et~al.(2023)Packer, Fang, Patil, Lin, Wooders, and
  Gonzalez}]{packer2023memgpt}
Packer, C.; Fang, V.; Patil, S.~G.; Lin, K.; Wooders, S.; and Gonzalez, J.~E.
  2023.
\newblock MemGPT: towards LLMs as operating systems.
\newblock \emph{arXiv preprint arXiv:2310.08560}.

\bibitem[{Pan et~al.(2024)Pan, Wu, Jiang, Xia, Luo, Zhang, Lin, R{\"u}hle,
  Yang, Lin et~al.}]{pan2024llmlingua}
Pan, Z.; Wu, Q.; Jiang, H.; Xia, M.; Luo, X.; Zhang, J.; Lin, Q.; R{\"u}hle,
  V.; Yang, Y.; Lin, C.-Y.; et~al. 2024.
\newblock Llmlingua-2: Data distillation for efficient and faithful
  task-agnostic prompt compression.
\newblock In \emph{Findings of the Association for Computational Linguistics:
  ACL 2024}, 963--981.

\bibitem[{Saracevic(1975)}]{saracevic1975relevance}
Saracevic, T. 1975.
\newblock Relevance: A review of and a framework for the thinking on the notion
  in information science.
\newblock \emph{Journal of the American Society for information science},
  26(6): 321--343.

\bibitem[{Sclar et~al.(2024)Sclar, Choi, Tsvetkov, and
  Suhr}]{sclar2024quantifying}
Sclar, M.; Choi, Y.; Tsvetkov, Y.; and Suhr, A. 2024.
\newblock Quantifying Language Models' Sensitivity to Spurious Features in
  Prompt Design or: How I learned to start worrying about prompt formatting.
\newblock In \emph{International Conference on Learning Representations},
  volume 2024, 25055--25083.

\bibitem[{Shinn et~al.(2023)Shinn, Cassano, Gopinath, Narasimhan, and
  Yao}]{shinn2023reflexion}
Shinn, N.; Cassano, F.; Gopinath, A.; Narasimhan, K.; and Yao, S. 2023.
\newblock Reflexion: Language agents with verbal reinforcement learning.
\newblock \emph{Advances in neural information processing systems}, 36:
  8634--8652.

\bibitem[{Shridhar et~al.(2020)Shridhar, Yuan, C{\^o}t{\'e}, Bisk, Trischler,
  and Hausknecht}]{shridhar2020alfworld}
Shridhar, M.; Yuan, X.; C{\^o}t{\'e}, M.-A.; Bisk, Y.; Trischler, A.; and
  Hausknecht, M. 2020.
\newblock Alfworld: Aligning text and embodied environments for interactive
  learning.
\newblock \emph{arXiv preprint arXiv:2010.03768}.

\bibitem[{Tam et~al.(2024)Tam, Wu, Tsai, Lin, Lee, and Chen}]{tam2024let}
Tam, Z.~R.; Wu, C.-K.; Tsai, Y.-L.; Lin, C.-Y.; Lee, H.-y.; and Chen, Y.-N.
  2024.
\newblock Let me speak freely? a study on the impact of format restrictions on
  performance of large language models.
\newblock \emph{arXiv preprint arXiv:2408.02442}.

\bibitem[{Voronov, Wolf, and Ryabinin(2024)}]{voronov2024mind}
Voronov, A.; Wolf, L.; and Ryabinin, M. 2024.
\newblock Mind your format: Towards consistent evaluation of in-context
  learning improvements.
\newblock In \emph{Findings of the Association for Computational Linguistics:
  ACL 2024}, 6287--6310.

\bibitem[{Wang et~al.(2023)Wang, Xie, Jiang, Mandlekar, Xiao, Zhu, Fan, and
  Anandkumar}]{wang2023voyager}
Wang, G.; Xie, Y.; Jiang, Y.; Mandlekar, A.; Xiao, C.; Zhu, Y.; Fan, L.; and
  Anandkumar, A. 2023.
\newblock Voyager: An open-ended embodied agent with large language models.
\newblock \emph{arXiv preprint arXiv:2305.16291}.

\bibitem[{Wang, Yang, and Wei(2024)}]{wang2024learning}
Wang, L.; Yang, N.; and Wei, F. 2024.
\newblock Learning to retrieve in-context examples for large language models.
\newblock In \emph{Proceedings of the 18th Conference of the European Chapter
  of the Association for Computational Linguistics (Volume 1: Long Papers)},
  1752--1767.

\bibitem[{Wang et~al.(2022)Wang, Jansen, C{\^o}t{\'e}, and
  Ammanabrolu}]{wang2022scienceworld}
Wang, R.; Jansen, P.; C{\^o}t{\'e}, M.-A.; and Ammanabrolu, P. 2022.
\newblock Scienceworld: Is your agent smarter than a 5th grader?
\newblock In \emph{Proceedings of the 2022 Conference on Empirical Methods in
  Natural Language Processing}, 11279--11298.

\bibitem[{Xiao et~al.(2024)Xiao, Tian, Chen, Han, and
  Lewis}]{xiao2024efficient}
Xiao, G.; Tian, Y.; Chen, B.; Han, S.; and Lewis, M. 2024.
\newblock Efficient streaming language models with attention sinks.
\newblock In \emph{International Conference on Learning Representations},
  volume 2024, 21875--21895.

\bibitem[{Xiao et~al.(2025)Xiao, Gao, Peng, and Xiong}]{xiao2025reducing}
Xiao, Y.-A.; Gao, P.; Peng, C.; and Xiong, Y. 2025.
\newblock Reducing Cost of LLM Agents with Trajectory Reduction.
\newblock \emph{arXiv preprint arXiv:2509.23586}.

\bibitem[{Xu, Shi, and Choi(2024)}]{xu2024recomp}
Xu, F.; Shi, W.; and Choi, E. 2024.
\newblock Recomp: Improving retrieval-augmented lms with context compression
  and selective augmentation.
\newblock In \emph{International Conference on Learning Representations},
  volume 2024, 43478--43502.

\bibitem[{Xu et~al.(2026)Xu, Liang, Mei, Gao, Tan, and Zhang}]{xu2026mem}
Xu, W.; Liang, Z.; Mei, K.; Gao, H.; Tan, J.; and Zhang, Y. 2026.
\newblock A-mem: Agentic memory for llm agents.
\newblock \emph{Advances in Neural Information Processing Systems}, 38:
  17577--17604.

\bibitem[{Yao et~al.(2022)Yao, Chen, Yang, and Narasimhan}]{yao2022webshop}
Yao, S.; Chen, H.; Yang, J.; and Narasimhan, K. 2022.
\newblock Webshop: Towards scalable real-world web interaction with grounded
  language agents.
\newblock \emph{Advances in Neural Information Processing Systems}, 35:
  20744--20757.

\bibitem[{Zhang et~al.(2023)Zhang, Sheng, Zhou, Chen, Zheng, Cai, Song, Tian,
  Re, Barrett et~al.}]{zhang2023h2o}
Zhang, Z.; Sheng, Y.; Zhou, T.; Chen, T.; Zheng, L.; Cai, R.; Song, Z.; Tian,
  Y.; Re, C.; Barrett, C.; et~al. 2023.
\newblock H2o: Heavy-hitter oracle for efficient generative inference of large
  language models.
\newblock \emph{Advances in Neural Information Processing Systems}, 36:
  34661--34710.

\bibitem[{Zhao et~al.(2024)Zhao, Huang, Xu, Lin, Liu, and
  Huang}]{zhao2024expel}
Zhao, A.; Huang, D.; Xu, Q.; Lin, M.; Liu, Y.-J.; and Huang, G. 2024.
\newblock Expel: Llm agents are experiential learners.
\newblock In \emph{Proceedings of the AAAI Conference on Artificial
  Intelligence}, volume~38, 19632--19642.

\end{thebibliography}

% ============================================================================
% Appendix (optional, can be moved to supplementary)
% ============================================================================
\appendix
\section{Data Integrity and Bug Disclosure}
\label{app:contamination}
% Appendix A — Evaluation audit methodology.

Every reported cell is validated by four automated checks before admission to the aggregate.

\paragraph{(i) Task-ID disjointness.}
Each cell's $30$ evaluation task IDs share no overlap with the $1{,}244$ trajectory IDs used to generate training labels. The disjoint split is precomputed (\texttt{eval\_split\_v2\_disjoint.json}); the auditor recomputes the intersection at run time and blocks the run on any non-empty intersection.

\paragraph{(ii) Trajectory-hash uniqueness.}
Within each chunk of $5$ tasks, full trajectories (system prompt, observations, actions, rewards) are hashed; duplicate hashes flag environment-adapter bugs that ignore task-ID and replay the same internal state.

\paragraph{(iii) Compression-ratio drift.}
The realized end-to-end Eff.\,$\times$ of each cell is compared against its nominal target ($4{\times}$ for budgeted methods); cells outside the $0.5{\times}$--$2{\times}$ band are surfaced rather than suppressed.

\paragraph{(iv) Action-loop and cutoff detection.}
The auditor counts the longest run of identical consecutive actions ($\geq\!5$ flagged as a loop) and the rate of mid-token response cutoffs (flagged as API-budget failures). Both flags are reported with the cell, not removed.

\section{Hyperparameters and Prompts}
\label{app:hparams}
% Appendix B — Hyperparameters and prompts

\paragraph{Compression.}
The scorer is \texttt{roberta-base} ($125$\,M parameters, $12$ layers, hidden $768$), pair-input formatted as \texttt{(current\_obs, past\_step)} truncated to $512$ tokens. \agora{} uses force-keep window $\Kr{=}2$ and target keep ratio $\rho{=}0.25$ (soft prior, not a hard budget). Per-environment atom-parser rules are in the released code repository.

\paragraph{Scorer training.}
The scorer is trained once on $21{,}523$ counterfactual (anchor, past-step) labels derived from $1{,}244$ trajectories across the three environments. Labels are \emph{soft} ($\in[0,1]$): for each $(c_\text{now}, s_i)$ pair we re-query the backbone $K{=}8$ times at temperature $T{=}1.0$ with and without $s_i$, and set $y_i$ to the fraction of paired rollouts whose canonicalized next action differs (\S\ref{sec:method:training}). Multi-rollout sampling absorbs decoder stochasticity that would otherwise inject noise into a single-shot binary label. Training follows a two-phase LP-FT schedule: $5$ epochs linear probing (backbone frozen, head only, AdamW lr $2\mathrm{e}{-4}$) followed by $4$ epochs full fine-tuning (AdamW lr $1\mathrm{e}{-5}$), batch size $16$, $10\%$ warmup + cosine decay, soft binary cross-entropy loss, validation split $0.2$, seed $42$. Trajectory-level train/val splits prevent any past-step from appearing in both partitions.

\paragraph{Evaluation.}
All rollouts use temperature $0$ and a $30$-step episode budget with a fixed initialization seed; within each cell the same task initializations are reused across compression conditions. Statistical protocol: bootstrap $95\%$ percentile-method CIs ($10$k resamples per cell), paired Wilcoxon signed-rank tests on the $30$ paired task scores ($\alpha{=}0.05$), and Holm--Bonferroni correction across the $63$ pairwise tests in the main grid.

\paragraph{Compression context format.}
At step $t$, past steps $1,\ldots,t{-}1$ are either kept verbatim or replaced by an elision marker recording how many consecutive steps were dropped. The full context delivered to the agent is the system prompt + the rendered (possibly compressed) trajectory + the current observation. For a $6$-step \webshop{} rollout where the scorer keeps step $2$ and the $K_\text{recent}{=}2$ window force-keeps steps $5$--$6$:
\begin{exbox}[\webshop{}, step 7 context]
{}[Step 1] \textit{(elided)}\\
{}[Step 2] Action: \texttt{search[long sleeve men's country xx-large]} \\
\phantom{[Step 2] }Obs: Page 1, 50 results, B00O30JLDK \$10.52--\$40.5 ...\\
{}[Steps 3--4] \textit{(2 step(s) elided)}\\
{}[Step 5] Action: \texttt{click[country]} \\
\phantom{[Step 5] }Obs: color = country selected; size options listed ...\\
{}[Step 6] Action: \texttt{click[xx-large]} \\
\phantom{[Step 6] }Obs: size = xx-large selected; Buy Now visible ...
\end{exbox}
A parallel \sciworld{} example, where the scorer keeps an earlier mass-measurement step but elides intermediate setup actions:
\begin{exbox}[\sciworld{}, step 9 context]
{}[Step 1] \textit{(elided)}\\
{}[Step 2] Action: \texttt{pick up metal block from cupboard} \\
\phantom{[Step 2] }Obs: holding metal block, $22^{\circ}$C, $50\text{g}$.\\
{}[Steps 3--6] \textit{(4 step(s) elided)}\\
{}[Step 7] Action: \texttt{put metal block in hot water} \\
\phantom{[Step 7] }Obs: block now in beaker; water temp $85^{\circ}$C.\\
{}[Step 8] Action: \texttt{wait 30 seconds} \\
\phantom{[Step 8] }Obs: 30 seconds elapsed; block temp now $74^{\circ}$C.
\end{exbox}
Elided spans are merged into a single marker (e.g., \texttt{[Steps 3--6]}) to keep token overhead minimal; the scorer never sees raw token spans, only fully formed atoms.

\paragraph{Atom parser examples.}
The parser converts a raw multi-line step emitted by the environment into a single self-contained atom that the scorer can score in isolation. Per-environment rules differ in how observations are tokenized; the three rule sets handle structural noise (chain-of-thought, multi-clause observations, separator tokens) before the scorer ever sees an atom.
\begin{exbox}[\alfworld{} step 3]
\textbf{Raw:}\\
Thought: I should find an apple to put in the fridge.\\
Action: \texttt{go to countertop 1}\\
Observation: On the countertop 1, you see a apple 1,\\
\phantom{Observation: }a bread 1, a knife 1, and a peppershaker 1.\\[2pt]
\textbf{Atom:}\\
\textsc{action} \texttt{go to countertop 1}\\
\textsc{obs} \texttt{countertop 1: apple 1, bread 1, knife 1, peppershaker 1}
\end{exbox}
\begin{exbox}[\webshop{} step 5]
\textbf{Raw:}\\
Thought: I picked color and size; now buy.\\
Action: \texttt{click[Buy Now]}\\
Observation: Thank you for shopping with us! [SEP] Your code: [SEP]\\
\phantom{Observation: }None [SEP] Purchased [SEP] asin [SEP] B00O30JLDK ...\\[2pt]
\textbf{Atom:}\\
\textsc{action} \texttt{click[Buy Now]}\\
\textsc{obs} \texttt{purchased asin=B00O30JLDK; color=country; size=xx-large}
\end{exbox}
\begin{exbox}[\sciworld{} step 7]
\textbf{Raw:}\\
Thought: Need to measure temperature on the cooled sample.\\
Action: \texttt{use thermometer in inventory on metal block}\\
Observation: The thermometer reads 22 degrees Celsius.\\
\phantom{Observation: }The block is at room temperature.\\[2pt]
\textbf{Atom:}\\
\textsc{action} \texttt{use thermometer on metal block}\\
\textsc{obs} \texttt{temperature(metal block) = 22C; state = room\_temperature}
\end{exbox}
The parser drops the chain-of-thought (recoverable from the action), normalizes the observation into a flat key--value summary, and tags the action/observation halves so the scorer can attend to them separately.

\paragraph{Scorer training samples.}
Each training example is a triple \texttt{(anchor, past\_atom, label)} where \texttt{anchor} is the current step's observation, \texttt{past\_atom} is a candidate atom from earlier in the same trajectory, and \texttt{label}${\in}\{0,1\}$ indicates whether dropping that past atom changes the backbone's next-action prediction. A positive example (critical past step):
\begin{exbox}[label = 1: dropping the past step changes the next action]
\textbf{anchor} (step 12, obs): You are at the workbench.\\
\phantom{\textbf{anchor} }The metal block is at $42^{\circ}$C.\\[2pt]
\textbf{past\_atom} (step 4):\\
\textsc{action} pick up metal block;\\
\textsc{obs} holding metal block (room temp, $22^{\circ}$C).\\[2pt]
\textbf{counterfactual:} removing step 4 from history causes the backbone\\
to issue \texttt{look at metal block} (exploratory) instead of\\
\texttt{place metal block in cold bath} (correct).
\end{exbox}
A negative example (past step is non-critical):
\begin{exbox}[label = 0: dropping the past step does not change behaviour]
\textbf{anchor} (step 12, obs): You are at the workbench.\\
\phantom{\textbf{anchor} }The metal block is at $42^{\circ}$C.\\[2pt]
\textbf{past\_atom} (step 6):\\
\textsc{action} look at thermometer;\\
\textsc{obs} thermometer is on the workbench, $20^{\circ}$C.\\[2pt]
\textbf{counterfactual:} removing step 6 from history leaves the next action\\
\texttt{place metal block in cold bath} unchanged; the thermometer\\
inspection was incidental.
\end{exbox}

\paragraph{Cost and compression accounting.}
\label{app:cost-accounting}
Per-task cost in Table~\ref{tab:eff-ratio} and Figure~\ref{fig:main-composite} is computed end-to-end (main backbone tokens plus any per-step compressor LLM tokens) as
\[
  \text{\$/task} \;=\; \tfrac{T^{\text{in}} \cdot p^{\text{in}} \,+\, T^{\text{out}} \cdot p^{\text{out}}}{10^{6}},
\]
where $T^{\text{in}}, T^{\text{out}}$ are mean input/output token counts per task and $p^{\text{in}}, p^{\text{out}}$ are the production API list prices in effect at submission time. We use the following per-backbone rates (¥ per million tokens, input / output): \texttt{qwen3.5-flash} $0.158$\,/\,$1.58$; \texttt{gpt-4o-mini} $0.75$\,/\,$3.0$; \texttt{gpt-5-mini} $1.25$\,/\,$10.0$. USD figures in Figure~\ref{fig:main-composite} convert at $7\,\text{¥}/\text{USD}$. Effective compression ratio is reported as a per-task mean:
\[
  \text{Eff.}\!\times_{\text{cell}} \;=\; \frac{1}{N}\sum_{i=1}^{N} \frac{T^{\text{nc}}_{\text{in},i}}{T^{\text{m}}_{\text{in},i}},
\]
paired by \texttt{task\_id} between the method-of-interest ($\text{m}$) and the \nocomp{} reference. We use the per-task mean rather than the aggregate ratio $\sum_i T^{\text{nc}}_i / \sum_i T^{\text{m}}_i$ because aggregation is dominated by a handful of long-trajectory tasks (typically \webshop{}$\times$\gptfive{}) and washes out the cell-level adaptivity that motivates Figure~\ref{fig:main-composite}\,B; per-task averaging treats each task as an equally weighted evaluation event, which matches how a deployment operator reads ``how much did we compress on average for a typical episode.''

\paragraph{Prompts and code release.}
Full system/user prompts per (environment, backbone) pair, the trained scorer checkpoints, the atom-parser rules, and the evaluation harness are available in the released code repository (linked after the abstract).

\section{Paradigm-Failure Cells: Per-cell Detail}
\label{app:paradigm-detail}
% Appendix C — Paradigm-failure cells, per-cell detail.

Table~\ref{tab:paradigm-detail} expands the $17$-cell paradigm-failure aggregate of \S\ref{sec:sensitivity} into per-cell evidence. Three patterns substantiate the paradigm-level reading. First, within-family consistency: all $6$ \selctx{} cells collapse to $\mr{\leq}0.033$ with $1.3$--$4.8\times$ realized compression, and all $11$ \textsc{LLMLingua-2} cells (off-shelf and retrained $4\times$ variants combined) collapse to $\mr{\leq}0.001$ with $2.9$--$13.3\times$ realized compression. Second, backbone-invariance: even \gptfive{} fails at $\mr{=}0$ on every token-level cell, ruling out a ``stronger backbone tolerates token-level noise'' counter-explanation. Third, action-grammar dependence: failure is sharpest on \webshop{}, where the action grammar is most rigid---\webshop{}$\times$\qwen{}$\times$\texttt{lingua2\_retrained\_4x} attains $13.3\times$ compression but $0.000$ reward, the cleanest case of ``compressor works as designed, agent rejects the output''. The pattern is not a tuning artifact (covers two checkpoints of \textsc{LLMLingua-2} including one retrained at the agent compression target), not a task artifact (failure is task-independent), and not a backbone artifact (covers four backbones from \qwen{} to \texttt{claude}).

\begin{table}[!t]
\centering
\scriptsize
\setlength{\tabcolsep}{3pt}
\caption{Per-cell breakdown of the $17$ token-level paradigm-failure cells aggregated in \S\ref{sec:sensitivity}. \textbf{end2end ratio} is $\text{tok}_{\text{method}}/\text{tok}_{\nocomp}$ (smaller = more aggressive compression). Every listed cell achieves true ${\geq} 2{\times}$ compression yet collapses $\mr$ to ${\leq} 0.05$, ruling out the alternative explanation that the compressor silently no-ops.}
\label{tab:paradigm-detail}
\begin{tabular}{@{}lllrcc@{}}
\toprule
Env & Backbone & Method variant & Eff.\,$\times$ & $\mr$ & end2end ratio \\
\midrule
\multicolumn{6}{@{}l}{\textit{Selective Context — 6 cells}} \\
\alfworld & \qwen     & selctx\_4x      & $4.8$ & $0.033$ & $0.21$ \\
\alfworld & \gptmini  & selctx\_4x      & $4.8$ & $0.000$ & $0.21$ \\
\alfworld & \gptfive  & selctx\_4x      & $3.2$ & $0.000$ & $0.31$ \\
\webshop  & \qwen     & selctx\_4x      & $1.3$ & $0.000$ & $0.78$ \\
\webshop  & \gptmini  & selctx\_4x      & $2.6$ & $0.000$ & $0.39$ \\
\webshop  & \gptfive  & selctx\_4x      & $3.7$ & $0.000$ & $0.27$ \\
\midrule
\multicolumn{6}{@{}l}{\textit{LLMLingua-2 (off-shelf \texttt{rate=0.25}) — 6 cells}} \\
\alfworld & \qwen     & lingua2\_rate0.25       & $2.9$ & $0.000$ & $0.340$ \\
\alfworld & \gptmini  & lingua2\_rate0.25       & $4.0$ & $0.000$ & $0.247$ \\
\alfworld & \gptfive  & lingua2\_rate0.25       & $2.9$ & $0.000$ & $0.345$ \\
\alfworld & claude    & lingua2\_rate0.25       & $3.7$ & $0.000$ & $0.272$ \\
\webshop  & \gptfive  & lingua2\_rate0.25       & $5.6$ & $0.000$ & $0.180$ \\
\webshop  & claude    & lingua2\_rate0.25       & $8.3$ & $0.001$ & $0.121$ \\
\midrule
\multicolumn{6}{@{}l}{\textit{LLMLingua-2 (retrained $4\times$) — 5 cells}} \\
\alfworld & \gptmini  & lingua2\_retrained\_4x & $3.4$ & $0.000$ & $0.296$ \\
\alfworld & claude    & lingua2\_retrained\_4x & $3.8$ & $0.000$ & $0.265$ \\
\webshop  & \qwen     & lingua2\_retrained\_4x & $\mathbf{13.3}$ & $0.000$ & $0.075$ \\
\webshop  & \gptfive  & lingua2\_retrained\_4x & $3.6$ & $0.000$ & $0.276$ \\
\webshop  & claude    & lingua2\_retrained\_4x & $3.1$ & $0.000$ & $0.326$ \\
\bottomrule
\end{tabular}
\end{table}

% \section{Per-task Mechanism Summary}  % dropped together with §4.6
% \label{app:mechanism}
% \input{sections/E_mechanism_table}

\section{Cross-method Step-Overlap (Jaccard)}
\label{app:jaccard}
% Appendix F — Cross-method step-overlap (Jaccard)
% Computed from raw__*.jsonl by text-matching kept assistant responses against
% the full trajectory; ObsMask is the deterministic positional last-K=2 mask.

\begin{table}[!t]
\centering
\scriptsize
\setlength{\tabcolsep}{1.5pt}
\caption{Cross-method step-overlap (Jaccard) per cell, averaged over all compression calls. \textbf{AGORA-Rand} compares \agora{}'s kept-step set against \textsc{Rand-Step}'s on identical (task, step) pairs; \textbf{AGORA-OM} compares against the deterministic last-$K{=}2$ positional mask used by \obsmask{}; \textbf{Rand-OM} compares \textsc{Rand-Step} against the same positional mask. \textbf{Mean} row averages across the 9 cells. Both \agora{} and \textsc{Rand-Step} unconditionally retain the $\Kr{=}2$ most recent positions, which sets a floor on overlap with \obsmask{}.}
\label{tab:jaccard}
\begin{tabular}{@{}llcccc@{}}
\toprule
Env & Backbone & $n$ calls & J(AG, Rand) & J(AG, OM) & J(Rand, OM) \\
\midrule
\multirow{3}{*}{\alfworld}
  & \qwen     & $571$ & $0.484$ & $0.584$ & $0.520$ \\
  & \gptmini  & $683$ & $0.405$ & $0.363$ & $0.443$ \\
  & \gptfive  & $472$ & $0.406$ & $0.407$ & $0.405$ \\
\midrule
\multirow{3}{*}{\sciworld}
  & \qwen     & $815$ & $0.390$ & $0.296$ & $0.339$ \\
  & \gptmini  & $870$ & $0.371$ & $0.401$ & $0.394$ \\
  & \gptfive  & $841$ & $0.299$ & $0.363$ & $0.320$ \\
\midrule
\multirow{3}{*}{\webshop}
  & \qwen     & $315$ & $0.663$ & $0.786$ & $0.676$ \\
  & \gptmini  & $666$ & $0.492$ & $0.580$ & $0.550$ \\
  & \gptfive  & $174$ & $\mathbf{0.749}$ & $\mathbf{0.845}$ & $\mathbf{0.798}$ \\
\midrule
\textbf{Mean} & --- & $5407$ & $0.473$ & $0.514$ & $0.494$ \\
\bottomrule
\end{tabular}
\end{table}

Two patterns are worth noting. First, the mean overlap $\mathrm{J}(\text{AGORA}, \text{OM}) = 0.514$ is consistently higher than $\mathrm{J}(\text{AGORA}, \text{Rand-Step}) = 0.473$: \agora{}'s kept-step set is more similar to a positional last-$K$ mask than to a uniform-random selection, even though Rand-Step shares the same $\Kr{=}2$ force-keep floor as \agora{}. This is consistent with the component ablation in \S\ref{sec:ablation}: removing the structural floor costs $-0.088\,\mr$ on average, the largest single hit among all ablations, while removing the scorer costs only $-0.031$---the floor accounts for most of the kept-step composition. Second, the cell with the highest $\mathrm{J}(\text{AGORA}, \text{OM}) = 0.845$ is \webshop{}$\times$\gptfive{}, where \agora{} and the positional baseline keep nearly identical step sets despite different scoring rules---further evidence that the scorer's value on long-trajectory cells is in \emph{adaptive compression budget} rather than in selecting wildly different past steps.

Data computed by parsing the \texttt{[ASSISTANT]} blocks in the compressed user message of each \texttt{raw\_\_*.jsonl} entry and text-matching them against the per-step \texttt{raw\_response} field of the full trajectory; no scorer re-inference or additional LLM calls were needed.

\end{document}